%% file: sample-acmtog-SIGGRAPH-submission.tex
\documentclass[acmtog, authorversion,nonacm]{acmart}

\usepackage{booktabs} % For formal tables
\usepackage{pifont}
\usepackage[linesnumbered,ruled,vlined]{algorithm2e}

\usepackage[ruled]{algorithm2e} % For algorithms

\SetAlFnt{\small}
\SetAlCapFnt{\small}
\SetAlCapNameFnt{\small}
\SetAlCapHSkip{0pt}

\acmJournal{TOG}
\acmYear{2025}
\acmMonth{12}

\acmDOI{10.1145/3757377.3763977}

\begin{document}

% Title portion
\title{Training-Free Instance-Aware 3D Scene Reconstruction and Diffusion-Based View Synthesis from Sparse Images}

\author{Jiatong Xia}
\affiliation{%
 \institution{The University of Adelaide}
 \country{Australia}
 \city{Adelaide}
 }
\email{jiatong.xia@adelaide.edu.au}

\author{Lingqiao Liu}
\authornote{Corresponding author.}
\affiliation{%
 \institution{The University of Adelaide}
 \country{Australia}
  \city{Adelaide}
 }
\email{lingqiao.liu@adelaide.edu.au}

\begin{abstract}
We introduce a novel, training-free system for reconstructing, understanding, and rendering 3D indoor scenes from a sparse set of unposed RGB images. Unlike traditional radiance field approaches that require dense views and per-scene optimization, our pipeline achieves high-fidelity results without any training or pose preprocessing. The system integrates three key innovations: (1) A robust point cloud reconstruction module that filters unreliable geometry using a warping-based anomaly removal strategy; (2) A warping-guided 2D-to-3D instance lifting mechanism that propagates 2D segmentation masks into a consistent, instance-aware 3D representation; and (3) A novel rendering approach that projects the point cloud into new views and refines the renderings with a 3D-aware diffusion model. Our method leverages the generative power of diffusion to compensate for missing geometry and enhances realism, especially under sparse input conditions. We further demonstrate that object-level scene editing—such as instance removal—can be naturally supported in our pipeline by modifying only the point cloud, enabling the synthesis of consistent, edited views without retraining. Our results establish a new direction for efficient, editable 3D content generation without relying on scene-specific optimization.
Project page: \href{https://jiatongxia.github.io/TID3R/}{https://jiatongxia.github.io/TID3R/}
\end{abstract}




\begin{teaserfigure}
\centering
\includegraphics[width=1\linewidth]{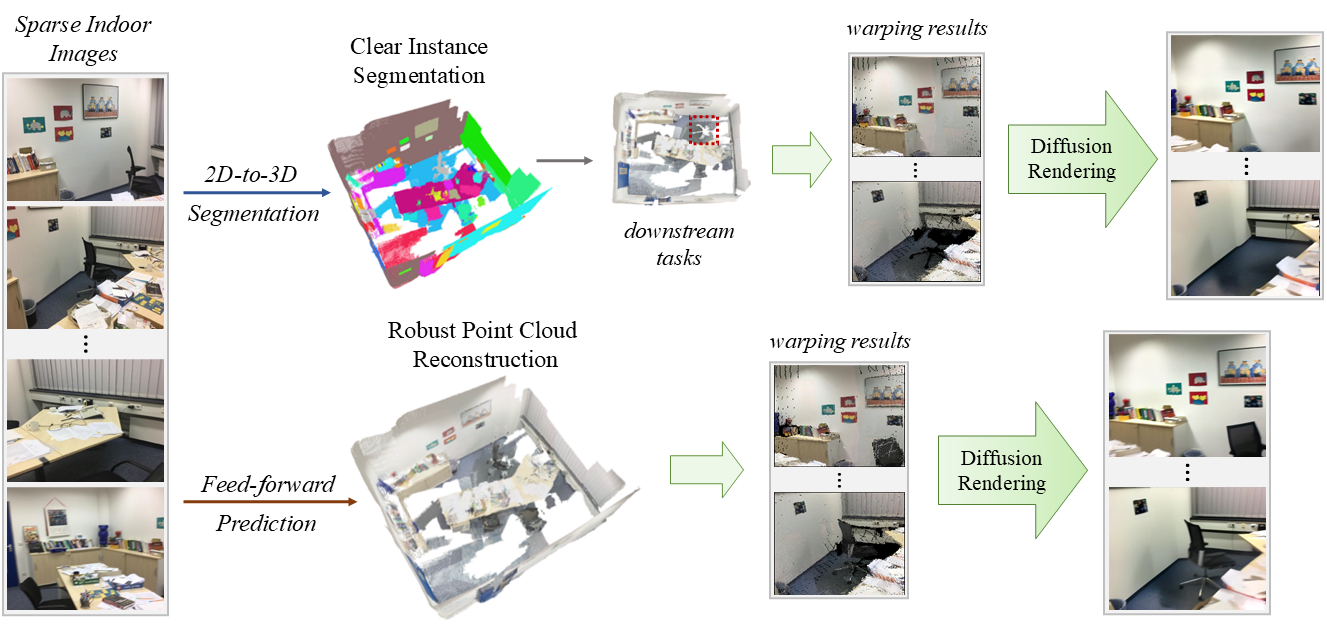} 
\caption{We present a novel, unified pipeline that transforms sparse image inputs into a clean, instance-aware point cloud without requiring any pose pre-processing or scene-specific learning, and can directly synthesizes photorealistic novel views through a tailored diffusion-based rendering approach. Beyond reconstruction and rendering, our method also showing good potential supports essential downstream tasks such as scene-level editing. 
}
\label{fig:teaser}
\end{teaserfigure}

\maketitle

\input{sec/1_introduction}
\input{sec/2_related}
\input{sec/3_method}
\input{sec/4_results}
\input{sec/5_conclusion}

% \clearpage

% Bibliography
\bibliographystyle{ACM-Reference-Format}
\bibliography{sample-bibliography}

\clearpage

\input{sec/appendix}

\end{document}

%% file: sec/1_introduction.tex
\section{Introduction}
\label{sec:intro}

Reconstructing a faithful 3D representation of a scene from a set of 2D source images—followed by rendering or simple editing of the reconstructed content—serves as a foundational workflow across various domains such as creative design, film production, game development, virtual reality (VR), and augmented reality (AR). While this pipeline has seen significant advances, it remains computationally demanding, particularly under sparse data regimes.

Radiance field-based approaches, such as NeRF~\cite{mildenhall2020nerf} and 3DGS~\cite{kerbl3Dgaussians}, have become dominant for achieving high-quality 3D reconstructions and novel view synthesis. These methods often yield photorealistic results and support texture or style editing when a sufficiently dense set of images is available. However, their reliance on scene-specific optimization and large input collections introduces substantial overhead. When input images are sparse or unposed--a common scenario in casual capture or lightweight applications--these methods struggle to maintain reconstruction fidelity or support interactive editing. Moreover, most existing techniques do not support structured manipulation at the object-instance level without additional fine-tuning. A training-free, sparse-view-capable system that provides high-fidelity rendering and semantic understanding remains largely absent.

To fill this gap, we propose a unified, training-free framework for reconstructing and rendering instance-aware 3D scenes using a sparse set of unposed RGB images. Our system eliminates the need for pose estimation or scene-specific training, enabling practical deployment in data-scarce or real-time settings. The core of our approach consists of three interconnected modules: geometry reconstruction, instance-aware perception, and novel-view rendering.

For the reconstruction stage, we build on MV-DUSt3R~\cite{tang2024mv} to predict dense 3D point clouds from unposed images in a purely feed-forward manner. MV-DUSt3R is a recent advancement in multi-view stereo reconstruction that leverages transformer-based architectures to process multiple views simultaneously, producing globally aligned point clouds without requiring camera calibration or pose estimation. While efficient, MV-DUSt3R outputs can still contain noisy or inconsistent geometry. To mitigate this, we introduce a warping-based anomaly elimination strategy that leverages multi-view depth consistency to prune unreliable points, resulting in robust 3D geometry reconstruction under sparse-view conditions.

To embed semantic understanding, we propose a warping-guided 2D-to-3D instance segmentation pipeline. Starting with 2D segmentation masks generated by a foundational model (e.g., SAM~\cite{kirillov2023segment}), we propagate these labels across views using estimated depth maps and warping flows. These are then lifted into a unified 3D point cloud, producing a view-consistent, instance-aware representation that supports downstream manipulation. Compared to prior methods such as PE3R~\cite{hu2025pe3r}, our approach achieves more accurate and coherent instance assignments across frames.

For rendering, we directly project the point cloud into a target view and apply a 3D-aware diffusion model, See3D~\cite{Ma2025See3D}, to refine the result. This design enables the model to leverage rich image priors to inpaint missing geometry, yielding photorealistic results even when the underlying point cloud is incomplete. Unlike classical methods that suffer from holes or artifacts under sparse input, our diffusion-guided approach can robustly synthesize plausible novel views. Although diffusion-based rendering does not currently achieve real-time performance, it is fast enough for many offline or interactive applications where quality is paramount.

Our framework also provides natural support for object-level scene editing. By modifying the instance-aware point cloud--such as removing a specific object instance--we can seamlessly produce consistent, edited novel views without retraining. This demonstrates the flexibility and extendability of our pipeline in handling basic scene modifications through simple, localized updates in the point cloud space.

In summary, our contributions span three major components, training-free dense reconstruction, instance-aware 2D-to-3D perception, and diffusion-guided rendering—offering an efficient and scalable solution to 3D scene modeling from sparse RGB inputs. Together, they establish a new direction for lightweight, semantically structured 3D scene understanding and rendering.

%% file: sec/2_related.tex
\section{Related Work}
\label{sec:related}

\begin{figure*}[!t]
\centering
\includegraphics[width=1\linewidth]{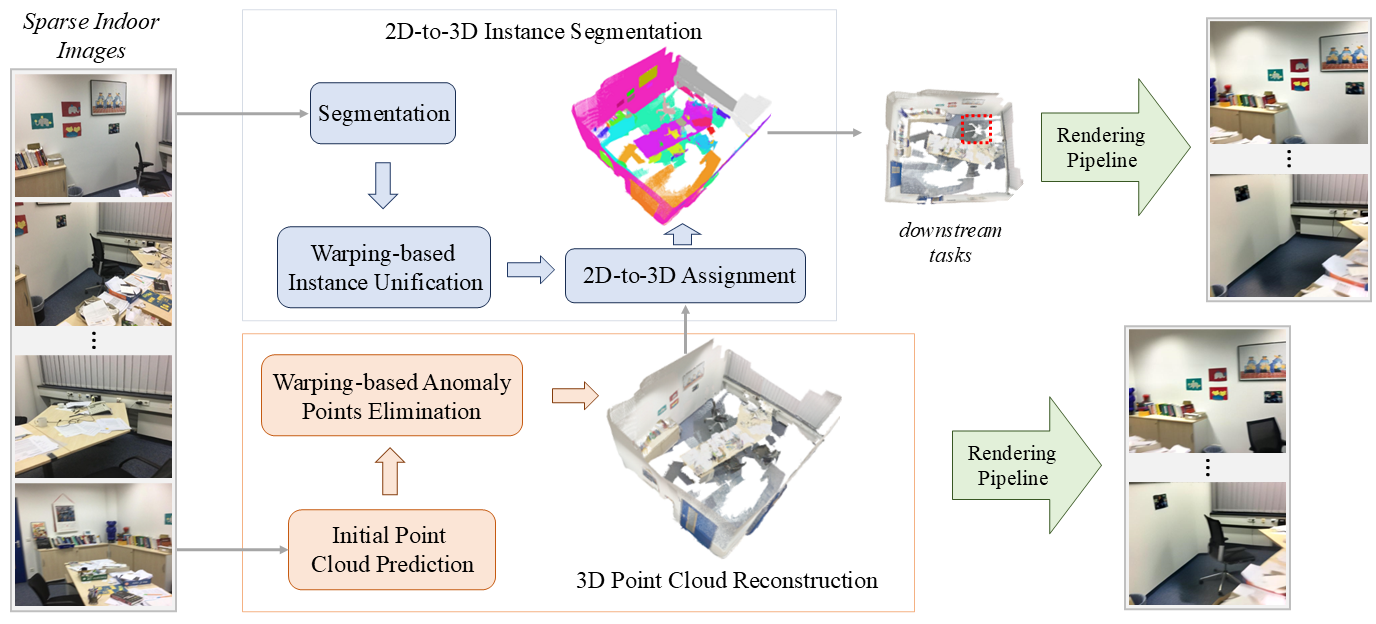} 
\caption{\textbf{Framework}.
We apply a feed-forward a feed-forward model (e.g., MV-DUSt3R) to predict the initial point cloud from the input unposed sparse images, followed by a novel warping-based anomaly points removal strategy to eliminate unreliable points in the initial point cloud, forming a clean and accurate scene point cloud. For 2D-to-3D instance segmentation, we first utilize a foundational segmentation model (e.g., SAM) to generate initial segmentation masks. We then introduce a warping-based instance unification strategy to align instances across frames. By associating the instance masks with the corresponding pointmaps, we obtain an instance-segmented 3D point cloud.
The instance-segmented 3D point cloud facilitates downstream tasks such as object-level editing. 
Finally, the point cloud can be directly rendered into high-quality 2D images through our designed diffusion-based rendering pipeline.
}
\label{fig:framework}
\end{figure*}

\subsection{3D Reconstruction from Images}
Recovering the geometry and appearance of a scene from a set of 2D images is a fundamental problem in computer vision and graphics. Traditional methods such as Structure-from-Motion (SfM)~\cite{schonberger2016structure} focus on reconstructing sparse 3D point clouds while jointly estimating camera parameters from a set of images. Based on estimated camera parameters, Multi-View Stereo (MVS) techniques~\cite{yao2018mvsnet,yu2020fast,yao2019recurrent} then aim to densely reconstruct the visible surfaces of the scene. Also, with estimated camera parameters and image sets, recent proposed radiance field methods such as NeRF~\cite{mildenhall2020nerf,barron2021mip,chen2022tensorf,barron2023zipnerf} and 3DGS~\cite{kerbl3Dgaussians,chen2024mvsplat,cheng2024gaussianpro,Yu2024MipSplatting,Huang2DGS2024,li2024dngaussian,zhao2024segs,zhang2024cor} enable high-fidelity 3D reconstruction and novel view synthesis, after scene-specific training. In parallel, numerous methods have been developed to directly infer 3D geometry from input images. Monocular depth estimation (MDE)~\cite{yang2024depth, yang2024depthany,bhat2023zoedepth,eigen2014depth}, for example, uses pre-trained networks to predict depth maps, which can generate pixel-aligned 3D point clouds using known camera parameters. More recently, DUSt3R~\cite{wang2024dust3r} and its extensions~\cite{yang2025fast3r,duisterhof2024mast3r, wang2025continuous} have introduced a novel approach that outputs per-image pointmaps, which can implicitly handle camera poses and recover dense 3D point clouds of every pixels from a set of unstructured input images. Building on this, MV-DUSt3R~\cite{tang2024mv} further enables the reconstruction of large-scale scenes from sparse multi-view images. Despite recent progress in point cloud reconstruction, a pipeline that enables direct rendering from point clouds to images is still missing, creating a bottleneck for broader 3D content creation.

\subsection{Diffusion Models}
Diffusion models~\cite{ho2020ddpm,song2020ddim} have emerged as a powerful class of generative models and have become the main approach to image generation~\cite{zhang2023adding} and video generation~\cite{ho2022video}, and were also extended to various domains, such as image restoration~\cite{yu2024supir}, and more~\cite{ke2024repurposing, duan2024diffusiondepth}. Further advances such as Score Distillation Sampling (SDS)~\cite{poole2022dreamfusion,tang2023dreamgaussian,liang2024luciddreamer} have also made it feasible to generate 3D assets. Moreover, diffusion-based methods~\cite{zhou2025stablecam,ziwen2025llrm,jiang2025rayzer,jin2025lvsm} have also been applied directly to the task of novel view synthesis. A recently proposed diffusion model See3D~\cite{Ma2025See3D}, proposes a reference-guided, warping-based generation framework. It utilizes known image information and warped pixel results of any selected viewpoints as input to restore the warped results, demonstrating strong performance in generating photorealistic image results that are strictly aligned with the projected pixel points. This capability makes See3D a promising tool for rendering pre-constructed 3D point clouds into high-fidelity 2D images.

\subsection{2D-to-3D Scene Perception}
Integrating state-of-the-art 2D foundation models~\cite{cheng2023deva,wang2022clip,oquab2023dinov2} such as SAM~\cite{kirillov2023segment} with 2D-to-3D reconstruction techniques has become one of the primary strategies for enabling scene-level 3D perception. Existing 2D-to-3D perception approaches mainly combined with NeRF~\cite{mildenhall2020nerf} or 3DGS~\cite{kerbl3Dgaussians}, they typically incorporate the features from 2D foundation models as additional attributes into the radiance field and learn them jointly, as in Feature3DGS~\cite{zhou2024feature}, GaussianGrouping~\cite{gaussian_grouping} and more~\cite{kerr2023lerf,cen2023segment}. Nevertheless, these radiance field methods still rely on scene-specific pretraining and are constrained by slow reconstruction processes. Recently, PE3R~\cite{hu2025pe3r} combined 2D perception models with DUSt3R~\cite{wang2024dust3r} to achieve feed-forward segmentation without requiring scene-specific pretraining. However, its reliance on the tracking model (e.g. SAM2~\cite{ravi2024sam}) limits its ability to maintain instance-consistent perception. Our proposed method better establishes consistent instance correspondences across multiple views, facilitates various downstream tasks such as scene editing.

\subsection{3D Scene Editing}
Recent progress in diffusion models has facilitated a surge of methods that combine them with NeRF~\cite{haque2023instruct,chen2024upst,fujiwara2024style} or 3DGS~\cite{wang2024view,xu2024texture,liu2024stylegaussian,chen2024mvip,huang20253dgic}, allowing for the editing of the radiance field. Compared with radience field or other 3D representations, point clouds offer a practical and flexible alternative, particularly suited for scene editing, as they are easier to obtain and more straightforward to manipulate. However, due to the absence of effective rendering methods for 3D point cloud, such approaches have not been fully explored. Our method first introduces a rendering pipeline tailored to this paradigm, facilitates instance-level scene editing based on reliable 3D point cloud reconstruction and instance perception, fills a gap in the field and offers a promising direction for future research.

%% file: sec/3_method.tex
\section{Method}
\label{sec:method}

\subsection{Problem Formulation}
Given a set of sparse images ${I_i}, I_i \in \mathbb{R}^{h \times w \times 3}, i \in [1, n]$ from the indoor scene, where each image has resolution $h \times w$ and $n$ is typically between 10 and 20, our objective is to reconstruct an accurate 3D representation of the scene that includes clearly segmented object instances and to synthesize novel views from arbitrary viewpoints. Achieving this goal presents several core challenges: (1) robustly reconstructing 3D geometry from sparse and unposed inputs, (2) generating instance-aware semantic understanding of the reconstructed scene, and (3) rendering visually coherent and photorealistic images from the resulting 3D content.

\paragraph{Framework}
Our proposed framework, illustrated in Fig.~\ref{fig:framework}, addresses these challenges through a unified, training-free pipeline. First, we adopt a feed-forward model to generate dense, pixel-aligned 3D point clouds from each input image, and introduce a novel anomaly removal strategy that leverages depth warping consistency across views to eliminate spurious or unreliable points. Second, we develop a 2D-to-3D instance segmentation module that uses warped depth maps and cross-view flow to propagate and unify 2D segmentation masks, enabling consistent instance labeling across views. The fused instance masks are then projected into 3D space, producing an instance-aware point cloud that supports downstream operations such as object removal. Finally, we propose a rendering pipeline tailored to sparse-view point clouds: it projects the 3D point cloud into target views and uses a 3D-aware diffusion model to inpaint missing regions and refine surface textures. This approach delivers high-quality renderings even under incomplete geometry, and proves robust in both reconstruction and post-editing scenarios.

\begin{figure}[ht]
\centering
\includegraphics[width=1\linewidth]{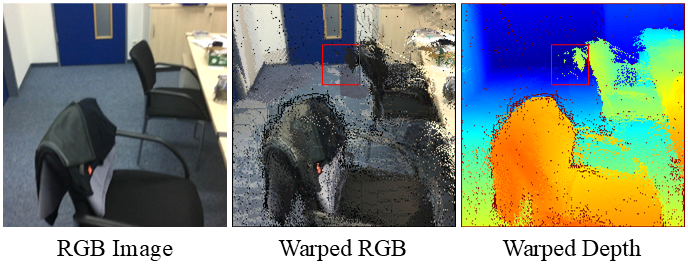} 
\caption{\textbf{Anomaly points}. The red box highlights anomaly points, which exhibit abnormal depth values.
}
\label{fig:noise}
\end{figure}

\subsection{Robust 3D Reconstruction from Images}
In practical scenarios, it is desirable to reconstruct 3D scenes from a small number of images, without relying on camera parameters, ground-truth depth, or any scene-specific training. MV-DUSt3R~\cite{tang2024mv} provides an effective solution to this problem by directly predicting scene geometry from sparse unposed views using a feed-forward architecture.

Specifically, given a set of input images ${I_i}, i \in [1, n]$, the model $\mathcal{D}$ predicts per-image depth maps ${Z_i}, Z_i \in \mathbb{R}^{h \times w}$, as well as the associated intrinsic and extrinsic camera parameters ${K_i}, K_i \in \mathbb{R}^{3 \times 3}$ and ${T_i}, T_i \in \mathbb{R}^{4 \times 4}$:
\begin{equation}
{Z_i}, {K_i}, {T_i} = \mathcal{D}({I_i}), \quad i \in [1, n].
\end{equation}
Using the depth map $Z_i$ and camera parameters $K_i, T_i$, we project each pixel in $I_i$ to its corresponding 3D position, forming a set of point clouds $\{\mathcal{P}_i\}$.
However, not all reconstructed points are reliable. MV-DUSt3R may produce outliers due to occlusions or estimation errors. To filter these, we propose a warping-based consistency check across views.

\noindent\paragraph{Warping-based anomaly point elimination} For each point in $\mathcal{P}i$, we compute its projection into a different image $j$ using the known transformation $T_j$ and camera matrix $K_j$, obtaining a warped depth map $W_{j}^i \in \mathbb{R}^{h \times w}$. In an ideal case, the warped depth $W_{j}^i(u, v)$ should be close to the native depth $Z_j(u, v)$, where $(u, v)$ is the projected coordinate in view $j$. 
If instead we observe
\begin{equation}
W_{j}^i(u, v) < \tau \cdot Z_j(u, v),
\end{equation}
for a threshold $\tau \in (0, 1)$, we infer that the point likely violates geometric consistency--e.g., it appears in front of the surface estimated in $j$ as shown in Fig~\ref{fig:noise}--and thus mark it as an outlier. This consistency check is applied across all pairs of views, and a binary mask $M^e_i(u, v)$ is constructed to record whether each point in $\mathcal{P}_i$ passes the consistency test.

Finally, we aggregate only the points that are consistently validated across views to form the final scene-level point cloud $\mathcal{P}_c$. This procedure effectively eliminates outliers and produces a clean, dense point cloud suitable for subsequent semantic segmentation and rendering.

\subsection{Instance Segmentation and Interactive Environment}
Accurate instance-level understanding of a 3D scene plays a crucial role in supporting downstream tasks such as semantic analysis and object-level editing. While existing 2D-to-3D segmentation approaches such as PE3R~\cite{hu2025pe3r} generate initial masks using SAM~\cite{kirillov2023segment} and propagate them across views using tracking models like SAM2~\cite{ravi2024sam}, they often struggle to maintain consistent instance assignments across views--particularly in sparse, room-scale environments. In contrast, our method achieves robust 2D-to-3D instance segmentation by leveraging geometric consistency via warping-based alignment, rather than relying on temporal tracking.

\paragraph{Warping-based instance unification}
We begin by applying a foundational segmentation model (e.g., SAM~\cite{kirillov2023segment}) to each input image $I_i$ to obtain a set of predicted instance masks ${\mathcal{S}^i_k}$, where $\mathcal{S}^i_k$ represents the $k$-th segmented region in frame $i$. For each mask $\mathcal{S}^i_k$, we retrieve the 3D point in the point cloud $\{\mathcal{P}_i\}$ that corresponds to the specified 2D pixel, forming a set of 3D point groups ${\mathcal{G}^i_k}$.

However, point groups from different images may correspond to the same physical object instance. To resolve this, we propose a fusion strategy based on geometric projection and mask matching. Specifically, for each point group $\mathcal{G}^i_k$, we project it onto the image plane of another view $j$ using camera parameters $K_j$ and $T_j$, resulting in a projected mask $\mathcal{S'}^j_k$. We then compare $\mathcal{S'}^j_k$ to all masks ${\mathcal{S}^j_l}$ in image $j$ using the mask Intersection over Union (mIoU):
\begin{equation}
\text{mIoU}(\mathcal{S'}^j_k, \mathcal{S}^j_l) > \eta,
\end{equation}
where $\eta$ is a predefined threshold. If the condition is satisfied, we consider the corresponding point groups $\mathcal{G}^i_k$ and $\mathcal{G}^j_l$ to represent the same instance and merge them into a unified group.
This process is repeated across all image pairs and point groups until no further merges can be performed. The result is a consolidated set of instance-level 3D point groups that are consistent across views.

\paragraph{Instance-level Interactive Environment}
Point clouds serve as a practical and versatile intermediate representation of 3D scenes in our framework. Compared to other representations such as meshes or radiance fields, point clouds are much easier to generate, refine and manipulate directly. When enhanced with instance segmentation, each object in the scene is naturally grouped as a collection of labeled 3D points, making the representation well-suited for targeted editing.

This structured design enables a variety of intuitive manipulation tools to be developed for instance-level editing. As a proof of concept, we demonstrate object removal by simply deleting all points associated with a given object instance from the reconstructed scene point cloud $\mathcal{P}_c$, yielding a modified version $\hat{\mathcal{P}}_c$ that reflects the editing operation.

To visualize the edited scene, we design a rendering pipeline that operates directly on the point cloud without relying on retraining or additional scene-specific tuning. Instead of converting the edited point cloud into another rendering format, we project it into the desired novel view and apply a 3D-aware diffusion model to inpaint and refine the image. This approach leverages the structure of the point cloud while taking advantage of strong image priors, enabling high-quality, photorealistic view synthesis under sparse input conditions.

\subsection{Point Cloud Rendering Pipeline}

\paragraph{Rendering with See3D}
While point clouds offer a convenient and editable 3D representation, they are inherently challenging to render into high-quality 2D images due to their sparse and discrete nature. Simply projecting the point cloud $\mathcal{P}_c$ onto a target view typically produces incomplete images with large unfilled regions caused by occlusions, sparse coverage, and non-uniform point distribution.

To address this, we incorporate a 3D-aware diffusion model, See3D~\cite{Ma2025See3D}, which refines such sparse projections to produce photorealistic images. See3D is a visual-conditional, multi-view diffusion model trained on large-scale Internet videos. It learns 3D priors directly from image sequences and operates without requiring explicit 3D supervision or camera pose annotations. In our framework, See3D utilizes user-provided input views as conditions to inpaint missing regions in a projected target view, enabling geometrically consistent and high-quality rendering.

Formally, for a given target viewpoint $T_k$, we first project the point cloud $\mathcal{P}_c$ onto its image plane, producing a sparse RGB image $\hat{I}_k \in \mathbb{R}^{h \times w \times 3}$, where many pixels may remain unfilled. We define a binary mask $M_k \in {0,1}^{h \times w}$ such that:
\begin{equation}
M_k(u, v) = 
\begin{cases}
1, & \text{if the pixel at } (u,v) \text{ is defined from projection} \\
0, & \text{otherwise}
\end{cases}
\end{equation}

Additionally, we use a set of original input images $\{I_i\}_{i=1}^{n}$ as reference views. The See3D model $\mathcal{M}$ takes the projected image $\hat{I}_k$, the corresponding mask $M_k$, and the reference set ${I_i}$ as input and synthesizes a completed image $\tilde{I}_k$ for the target view:

\begin{align}
    \tilde{I}_k = \mathcal{M}(\hat{I}_k, M_k, \{I_i\}_{i=1}^n),
\end{align}where $\tilde{I}_k \in \mathbb{R}^{h \times w \times 3}$ is the final rendered image.

This rendering pipeline enables us to directly generate photorealistic images from point cloud representations without any scene-specific optimization or retraining, supporting a fully training-free and flexible reconstruction-to-rendering workflow.

\paragraph{Rendering for Edited Point Clouds}
When the point cloud $\mathcal{P}_c$ undergoes editing operations--such as object removal--the geometric structure of the scene changes. Consequently, the original input images $\{I_i\}_{i=1}^n$ may no longer be fully consistent with the modified scene, especially in regions corresponding to the removed content. Using these unaltered images as references can introduce inconsistencies and artifacts in the synthesized views.

To address this, we apply a reference masking strategy that excludes affected regions in the source images. Specifically, we identify these regions by projecting the removed portions of the edited point cloud onto each input image plane. The corresponding pixels in the original image $I_i$ are then set to zero, producing a masked reference image $I_i$ that retains only the unaltered content. This results in a set of filtered reference images $\{I_i\}_{i=1}^n$, which serve as reliable and scene-consistent conditioning inputs for the See3D model.

Empirically, we find that this masking strategy significantly improves rendering quality by ensuring geometric consistency with the modified scene and eliminating contradictions that may arise from using outdated or misaligned references. Importantly, this approach supports efficient, high-fidelity synthesis of edited views without requiring additional training or fine-tuning of the model.

%% file: sec/4_results.tex
\section{Results}
\label{sec:results}
In this section, we present our experimental setup and results. In Sec.~\ref{sec:implement} we provide a detailed description of the implementation details, including experimental setup, dataset and evaluation metrics, and the procedure for rendering from specified test viewpoints. In Sec.~\ref{sec:main_results}, we showcase the main results of our method. Finally, in Sec.~\ref{sec:ablation}, we conduct ablation studies to evaluate the contributions of different components of our proposed framework.

\subsection{Implementation}
\label{sec:implement}

\paragraph{Experimental details.}
In our framework, we adopt MV-DUSt3R~\cite{tang2024mv} to obtain the initial feed-forward point cloud, and apply our anomaly point elimination methods with the parameters $\tau$ set to $\frac{3}{4} $. To generate the initial segmentation masks, we adopt MobileSAMv2~\cite{zhang2023mobilesamv2}, a lightweight variant of SAM~\cite{cen2023segment}, and in our unification process, we set the threshold $\eta$ to $\frac{1}{3} $. In our diffusion rendering pipeline, we use the pretrained See3D~\cite{Ma2025See3D} model to render the point cloud. We implement other comparison methods in our experiments using their source code. We conduct our experiments on NVIDIA 4090 GPUs.

\paragraph{Datasets and evaluation metrics.}
We first conduct instance segmentation experiments on five selected scenes with sparse input images from the ScanNet~\cite{dai2017scannet} dataset to evaluate the performance of our proposed 2D-to-3D instance segmentation pipeline, we report quantitative results using standard metrics including AP, AP50, and AP75 to assess accuracy between the ground truth label and our matched instance segmentation mask. Following previous work such as CUSt3R~\cite{wang2025continuous}, we perform video depth evaluation to assess both per-frame depth accuracy and inter-frame depth consistency of each method. We report Root Mean Square Error (RMSE) and $\delta$  < 1.25 as our evaluation metrics. In addition, we evaluate the novel-view image synthesis quality of our rendering pipeline using PSNR, SSIM, and LPIPS. We conduct these evaluations across depth estimation, rendering quality on a set of seven selected scenes with sparse images from ScanNet++~\cite{yeshwanth2023scannet++} and ScanNet~\cite{dai2017scannet} testing sets, and present the overall results.

\paragraph{Testing novel view poses.}
Our method enables reconstruction and rendering without requiring input camera parameters. In the absence of input camera poses, rendering from a given test viewpoint necessitates an additional camera pose alignment step. Specifically, we compute the transformation between the ground-truth poses of the input images and the corresponding predicted poses from the feed-forward reconstruction network. This process yields a rotation and translation matrix that maps coordinates from the ground-truth pose space to the predicted pose space. We then apply the same transformation to the test image's ground-truth pose to obtain its corresponding pose in the predicted coordinate system, enable given camera poses testing.

\subsection{Main Results}
\label{sec:main_results}
In this part, we present the main experimental results. We begin with the instance segmentation results to demonstrate the effectiveness of our proposed 2D-to-3D instance segmentation method, followed by the depth estimation results, which evaluate the performance of our reconstruction pipeline. Finally, we show the rendering results to assess the quality of our rendering pipeline under novel viewpoints.

\begin{figure*}[p]
\centering
\includegraphics[width=0.98\linewidth]{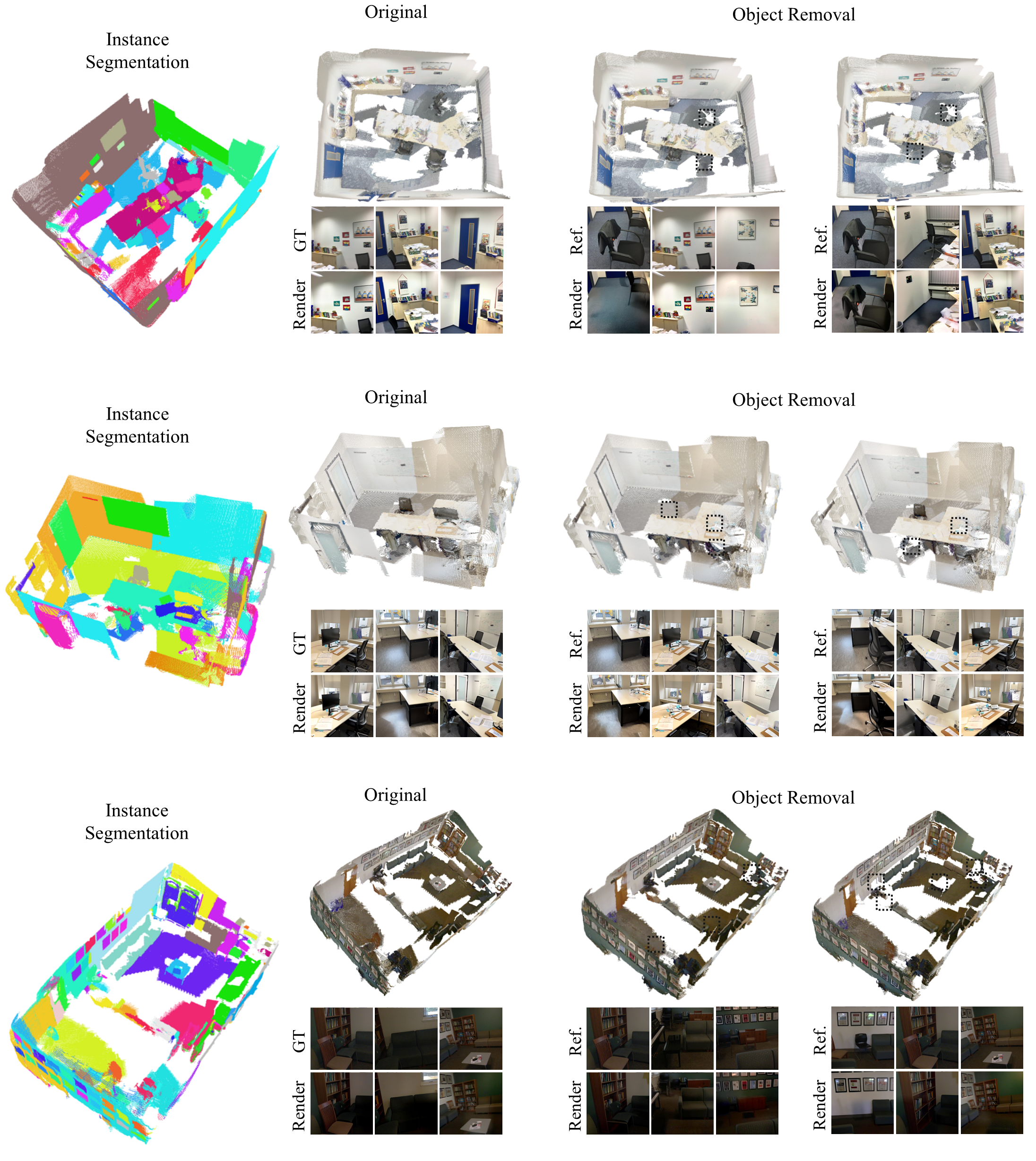} 
\caption{\textbf{Main results}. We present the main experimental results, including instance segmentation, point cloud reconstruction, novel view synthesis, and scene edits through object removal. Each row corresponds to a test scene. The left column shows the instance segmentation results. The center section labeled ``Original'' displays the reconstructed point cloud (top) and novel view synthesis results (bottom, with ground truth images on the top and rendered results on the bottom) of original input images. The two columns on the right illustrate two different object removal results, with the removed objects are highlighted with black dashed boxes in the point clouds
}
\label{fig:main_results}
\end{figure*}

\begin{figure*}[!t]
\centering
\includegraphics[width=0.92\linewidth]{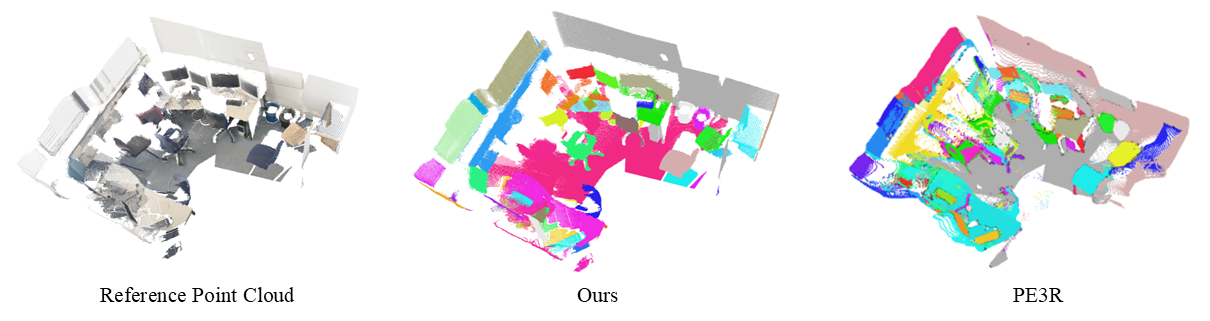} 
\vspace{-3mm}
\caption{\textbf{Instance segmentation}. We present a visual comparison between our 2D-to-3D instance segmentation results and those of PE3R, with the reference raw 3D point cloud on the left. While PE3R's reconstructed point cloud appears to be of lower quality than ours, it is worth noting that their segmentation results are produced independently and are not affected by the reconstruction quality.
}
\label{fig:seg}
\end{figure*}

\paragraph{Instance segmentation}
In Fig.~\ref{fig:main_results}, we show the 3D instance segmentation visualization of the reconstructed scene on the left side. As can be seen, our method achieves a clear and distinct instance segmentation, where different objects are easily distinguishable by color. To quantitatively evaluate our performance, we conduct experiments on ScanNet scenes with ground-truth instance segmentation labels. As shown in Tab.~\ref{tab:instance}, our method significantly outperforms other 2D video frames or the 2D-to-3D segmentation method, with a substantial improvement in overall AP. A more intuitive comparison is provided in Fig.~\ref{fig:seg}, where our method produces well-separated individual instances. Overall, our approach demonstrates remarkable performance in 2D-to-3D point cloud instance segmentation.

\begin{table}[t]
    \caption{\textbf{Instance segmentation evaluation}. We show overall instance segmentation performance using AP, AP50, and AP75 metrics on the selected scenes.
    }
    \label{tab:instance}
    \centering
    \resizebox{0.82\columnwidth}{!}{
    \begin{tabular}{c c c c}
        \toprule
         Method & AP ↑ &  AP50 ↑ & AP75 ↑   \\
        \midrule
        DEVA~\cite{cheng2023deva}      & 13.9  & 25.9 & 11.7   \\
         PE3R~\cite{hu2025pe3r}        & 20.2  & 32.6 & 22.3 \\
         w/o Warping-based unifi.    & 17.0  & 26.8 & 18.1 \\
         \textbf{Ours}       &  \textbf{25.0} & \textbf{41.8} & \textbf{25.4} \\
        \bottomrule
    \end{tabular}}
\end{table}

\paragraph{Reconstruction}
In Fig.~\ref{fig:main_results}, we present the reconstruction results of our method across various scenes. As shown in the visualized point clouds, our approach achieves clean and high quality reconstructions without anomaly points. Quantitative depth evaluation results in Tab.~\ref{tab:reconstruction} further demonstrates our reliable reconstruction, where our method outperforms existing approaches, achieving better depth accuracy. This provides a solid foundation for subsequent point cloud warping and the diffusion-based rendering pipeline.

\begin{table}[t]
    \caption{\textbf{Depth evaluation}. We show the overall video depth evaluation results on selected reconstruction scenes.
    }
    \label{tab:reconstruction}
    \centering
    \resizebox{0.89\columnwidth}{!}{
    \begin{tabular}{c c c }
        \toprule
         Method &    RMSE ↓ & $\delta <1.25$ ↑    \\
        \midrule
         DepthAnythingV2~\cite{yang2024depthany}        &  0.577 & 0.689   \\
         PE3R~\cite{hu2025pe3r}        &  0.378 & 0.905   \\
         MV-DUSt3R+~\cite{tang2024mv}  &  0.219 & 0.977    \\
         \textbf{Ours}       &  \textbf{0.209} & \textbf{0.981}   \\
        \bottomrule
    \end{tabular}}
\end{table}

\begin{figure*}[!t]
\centering
\includegraphics[width=1\linewidth]{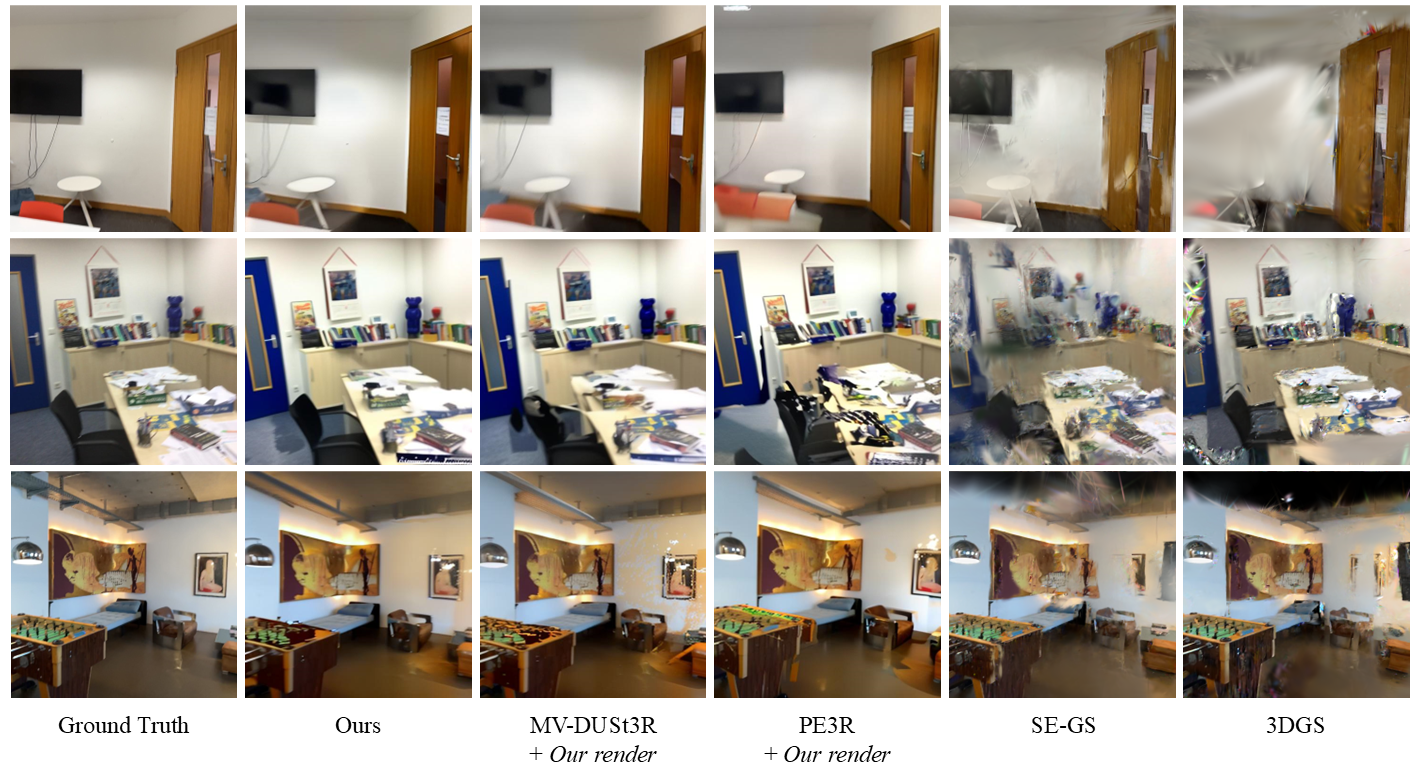}
\caption{\textbf{Novel views rendering}. We show the visualization comparisons on novel view synthesis.
}
\label{fig:render}
\end{figure*}

\begin{table*}[t]
    \caption{\textbf{Rendering evaluation on novel views}. We show the overall novel views synthesis performance of methods on selected reconstruction scenes. We indicate whether each method requires camera parameter input or per-scene training. We apply our rendering pipeline to the reconstructed point clouds produced by PE3R and MV-DUSt3R for a comparison.
    }
    \label{tab:render}
    \centering
    \resizebox{1.5\columnwidth}{!}{
    \begin{tabular}{c c c c c c c}
        \toprule
         & Poses & Training~ &\multicolumn{3}{c}{Novel View Synthesis}  \\
         Method & ~~~~~~~~~Free~~~~~~~~  & ~~~~~~~~~Free~~~~~~~~  &  ~~~PSNR ↑~~~  & ~~~~~SSIM ↑~ & ~LPIPS ↓~  \\
        \midrule
        3DGS~\cite{kerbl3Dgaussians}   &\ding{55}  &\ding{55}    & 15.61  & 0.602 & 0.541   \\
        2DGS~\cite{Huang2DGS2024}   &\ding{55}  &\ding{55}    & 15.55  & 0.591 &  0.544   \\
        GaussianGroup.~\cite{gaussian_grouping}   &\ding{55}  &\ding{55}    & 15.72  & 0.610 & 0.532   \\
        DNGaussian~\cite{li2024dngaussian}   &\ding{55}  &\ding{55}    & 16.82 & 0.627 &  0.549   \\
        CoR-GS~\cite{zhang2024cor}   &\ding{55}  &\ding{55}    &  17.12 & 0.631 & 0.496    \\ 
        SE-GS~\cite{zhao2024segs}   &\ding{55}  &\ding{55}    &  17.03 & 0.635 & 0.502    \\ 
        LVSM~\cite{jin2025lvsm}   &\ding{55}  &\ding{51}    & 12.92  & 0.617 &  0.620   \\
         PE3R~\cite{hu2025pe3r} + \textit{Our render.}~     & \ding{51} & \ding{51}   & 14.68  & 0.560 &  0.529  \\
         MV-DUSt3R+~\cite{tang2024mv} + \textit{Our render.}~ & \ding{51} & \ding{51}   & 17.69 & 0.613 & 0.458  \\
         \textbf{Ours}       & \ding{51} & \ding{51}  &  \textbf{18.15} & \textbf{0.645} & \textbf{0.434}   \\
        \bottomrule
    \end{tabular}}
    \vspace{5pt}
\end{table*}

\paragraph{Rendering}
We evaluate the rendering performance of our method on novel views for each reconstructed scene under sparse-view inputs. As shown in Fig.~\ref{fig:main_results}, our approach achieves high-quality rendering results from unseen viewpoints. The rendered images exhibit clear visual details and strong consistency with the ground truth images, demonstrating the high fidelity of our rendering pipeline. Furthermore, we provide visualizations of object removal rendering results to demonstrate the capability of our rendering pipeline in handling edited scenes. Within the same reconstructed scene, after removing selected objects, our method is able to generate geometrically consistent and visually sharp images without requiring any additional input, showcasing the effectiveness and generalization ability of our approach in supporting scene-level editing.

In Fig.~\ref{fig:render} and Tab.~\ref{tab:render}, we compare our approach with existing methods under novel views in the reconstructed scenes. As shown in Tab~\ref{tab:render}, our method achieves a superior performance compared to all approaches in a training-free manner and without requiring camera pose inputs. Notably, our method exhibits advantages over 3DGS-based approaches that are explicitly tailored for sparse input views, such methods are generally designed for forward-facing capture settings or relatively simple 360-degree object-centric scenarios, and thus tend to deteriorate significantly when applied to room-scale scenes with limited viewpoints and failing to produce satisfactory reconstructions. In contrast, our results show a notable improvement in visual quality in Fig.~\ref{fig:render}, highlighting the robustness and effectiveness of our pipeline.
Diffusion-based novel views rendering pipeline LVSM generates novel views by interpolating video frames between input images, and failed in the view generation due to the severely insufficient overlap between input frames in this case. We also employ our rendering pipeline to generate novel views from the point clouds reconstructed by PE3R and MV-DUSt3R. Benefiting from our higher-quality reconstructions, our rendering results also consistently outperform those methods, demonstrating the effectiveness of our unified pipeline. 

\begin{figure*}[!t]
\centering
\includegraphics[width=0.74\linewidth]{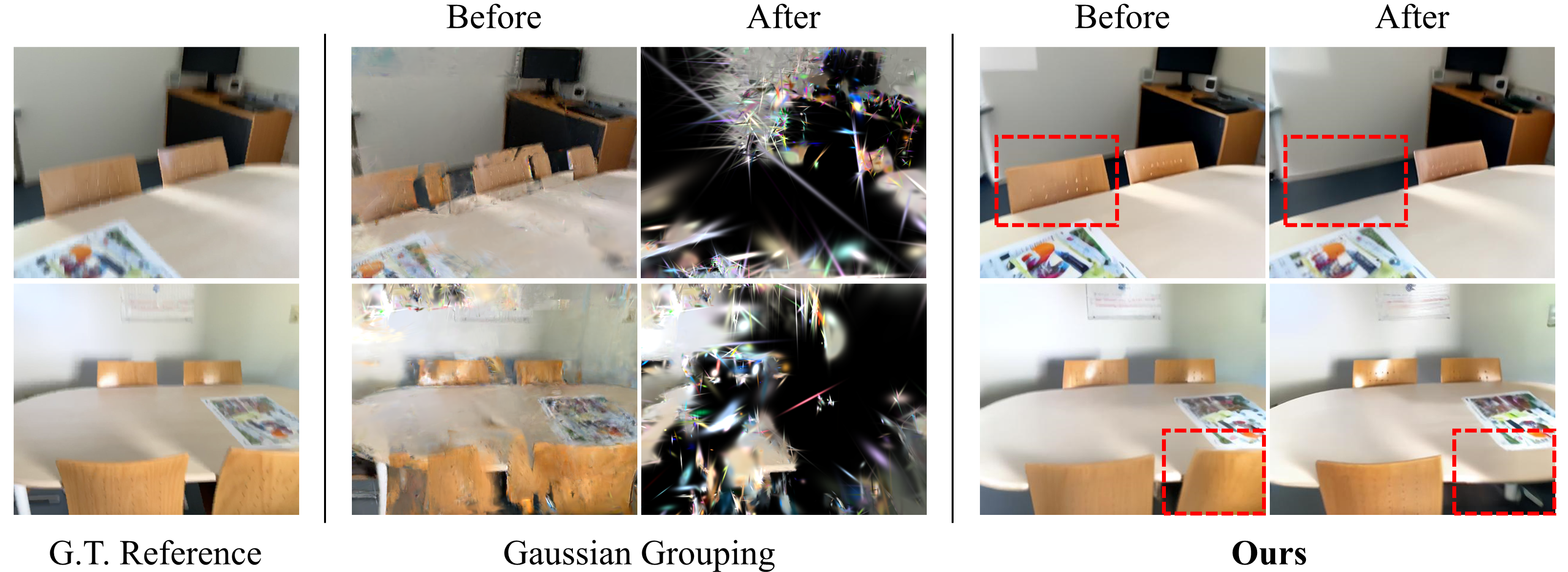} 
\caption{\textbf{Object removal comparison}. We present the visualization results of removing the same object under two different views, with the original ground-truth image on the left as reference. For each method, we display the rendered results before and after object removal. The removed object is highlighted with a red box in our results.
}
\label{fig:edit_comp}
\end{figure*}

In Fig.~\ref{fig:edit_comp}, we show a visual comparison of the object removal results between our method and Gaussian-Grouping. Due to the failure of Gaussian Splatting reconstruction under sparse-view conditions, Gaussian-Grouping erroneously removes an excessive number of Gaussians during object removal, resulting in large unfilled holes in the final output. In contrast, our method achieves superior rendering quality both before and after object removal, and produces clear and consistent results that align well with the intended editing objective.

\begin{table}[t]
    \caption{\textbf{Runtime costs}. We show a breakdown of processing cost comparison including camera poses processing, scene reconstruction and views inference between our overall pipeline and conventional approaches.
    }
    \label{tab:time}
    \centering
    \resizebox{1\columnwidth}{!}{
    \begin{tabular}{c c c c c}
        \toprule
         Method & Poses &  Reconstruct & Inference  &  Total \\
        \midrule
         NeRF~\cite{mildenhall2020nerf}        & 2 mins  & 146 mins & 1.5 mins & 149.5 mins \\
        3DGS~\cite{kerbl3Dgaussians}        & 2 mins  & 7 mins & - & 9 mins \\
         \textbf{Ours}       &  \textbf{-} & \textbf{1 min} & 2.5 mins & \textbf{3.5 mins} \\
        \bottomrule
    \end{tabular}}
\end{table}

\paragraph{Runtime}
Tab.~\ref{tab:time} shows a detailed overall time costs under the same setting of 20 input images and 10 test images. Compared to conventional approaches, our method eliminates the need for camera pose processing while achieving substantial advantages in the reconstruction stage and maintaining competitive overall performance.

\subsection{Ablation Studies}
\label{sec:ablation}
In this section, we conduct ablation studies to evaluate the effectiveness of our proposed components across instance segmentation, reconstruction, and rendering pipelines.

\paragraph{Instance unification.}
Without employing the warping-based instance unification strategy, the initial masks from individual 2D frames are directly projected into the 3D space without establishing inter-frame consistency, which could be viewed as a lower bound of the 2D-to-3D instance segmentation performance in the same setting. As shown in Tab.~\ref{tab:instance}, our method improves AP50 by 15 points compared with the results without warping-based instance unification, validates the effectiveness of our proposed component.

\begin{figure*}[!t]
\centering
\includegraphics[width=1\linewidth]{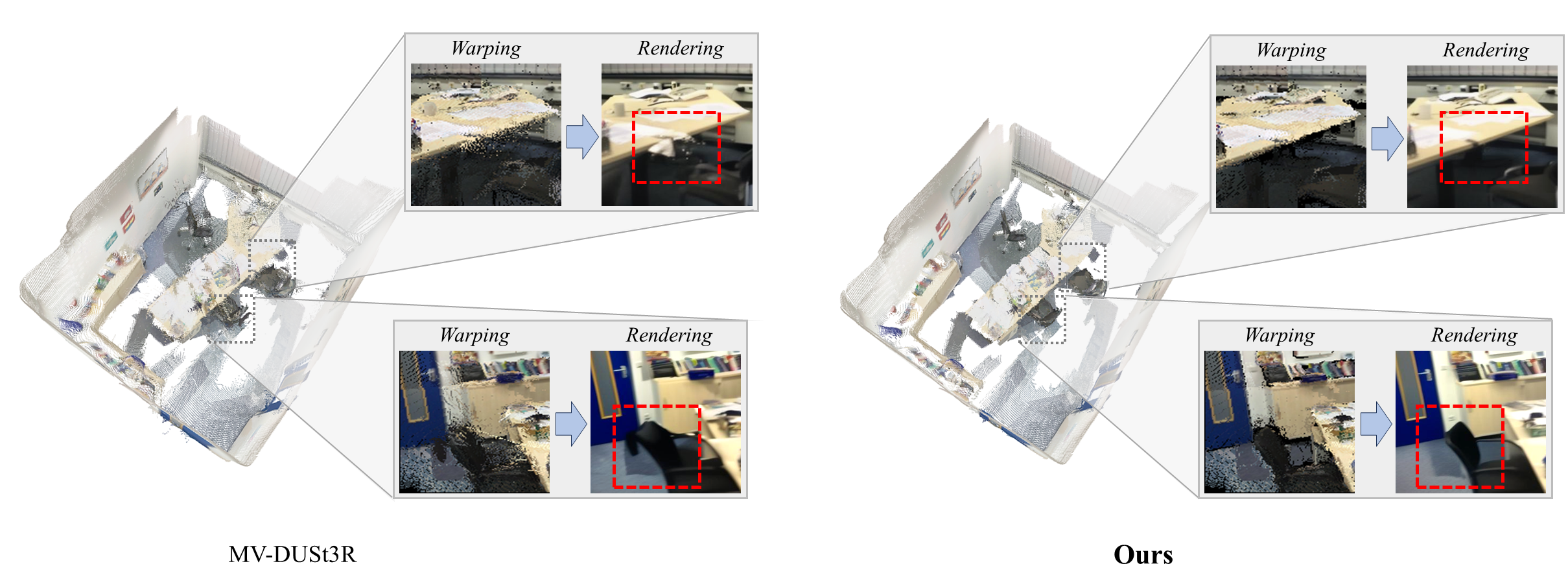} 
\caption{\textbf{Anomaly points elimination}. We show the reconstruction, warping, and rendering visualization results before and after anomaly point elimination. On the left, the baseline method without anomaly point removal shows a large number of outlier points in the reconstructed point cloud, which significantly degrades the warping and final rendering results. In contrast, our proposed method effectively eliminates these anomaly points, resulting in cleaner reconstructions and better rendering quality.
}
\label{fig:points}
\end{figure*}

\paragraph{Anomaly points elimination.}
As shown in Fig.~\ref{fig:points}, the baseline reconstruction result suffers from anomaly points, which introduce noise on the 2D warping results and adversely affect subsequent rendering. In contrast, our warping-based anomaly elimination effectively removes these outliers, resulting in cleaner and more consistent renderings. This improvement is also evident in the depth comparisons presented in Tab.~\ref{tab:reconstruction}, where our proposed component leads to better reconstruction quality compared to the baseline.

\begin{figure*}[!t]
\centering
\includegraphics[width=0.96\linewidth]{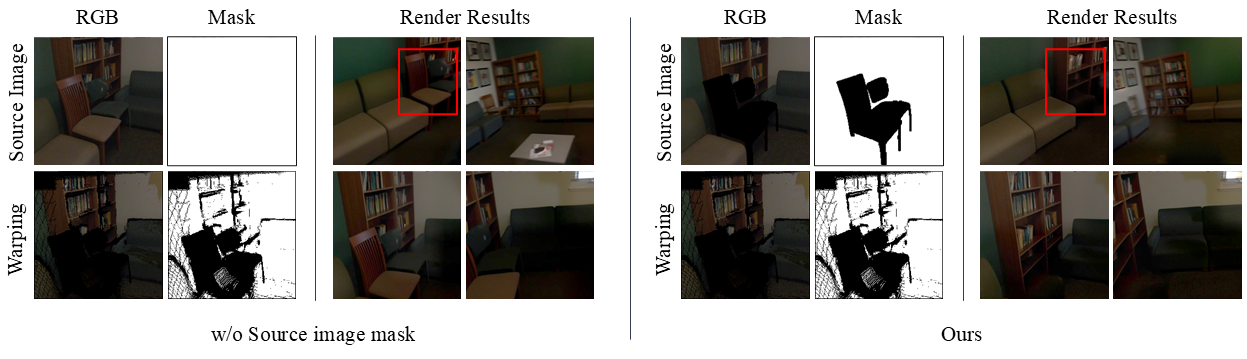} 
\caption{\textbf{Rendering masks}.We show the object removal results with and without applying rendering masks to the source images. Without rendering masks, the rendered outputs tend to preserve the full geometry from the source images, even after the object has been removed from the point cloud. This leads to a failure in reflecting the intended object removal in the final rendered image.
}
\label{fig:ref_mask}
\end{figure*}

\paragraph{Rendering masks.}
Integrating valid masks for source images into our diffusion rendering pipeline significantly enhances its compatibility with edited scenes. To verify the effectiveness of the source image masks, we compare rendering results with and without using them in Fig.~\ref{fig:ref_mask}. As we can clearly observe, without the masks, the final renderings of the warped results from the modified point cloud remain affected by geometric inconsistencies in the source images. In contrast, our method produces geometrically consistent and visually sharp images, clearly demonstrating the effectiveness of the proposed component.

%% file: sec/5_conclusion.tex
\section{Conclusion}
\label{sec:conclusion}
In this work, we presented a novel training-free pipeline for 3D reconstruction, understanding, and rendering of indoor scenes from sparse, unposed RGB images, offering a more flexible and efficient alternative to conventional radiance field techniques. Central to our approach are three innovations: a warping-based point cloud filtering mechanism for robust geometry reconstruction, a warping-guided 2D-to-3D instance lifting strategy for semantic scene understanding, and a 3D-aware diffusion-based rendering pipeline that achieves high-quality image renderings. Notably, our proposed framework also supports intuitive object-level editing through simple point cloud modifications. These contributions open new avenues for interactive and generalizable 3D content generation, bridging the gap between geometry-based methods and generative models. 

%% file: sec/appendix.tex
\appendix

\section{Editing Showcase}
\begin{figure}[t]
\centering
\includegraphics[width=1\linewidth]{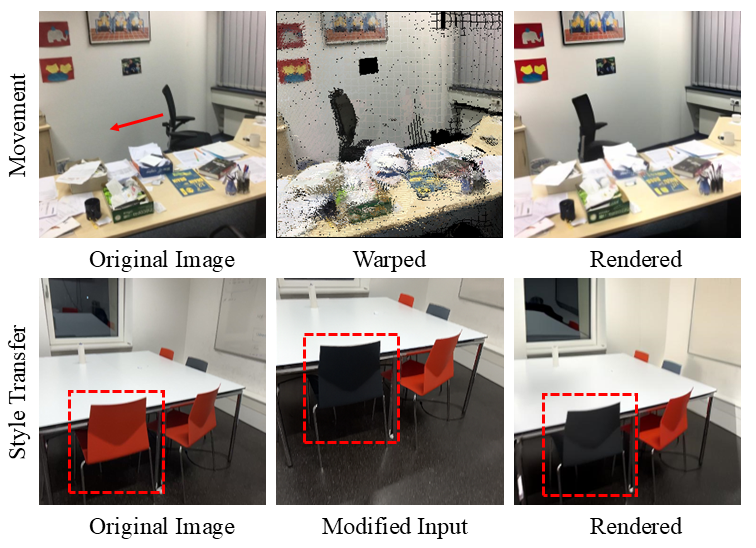} 
\caption{\textbf{Editing tasks example}. We show the original image of the view on the left. We apply a object point-cloud translation (red arrow) in the first row, followed by warping and rendering. We modify the object (red box) color of the input images before reconstruction, and the final rendered results reflect the applied color changes.
}
\label{fig:edit}
\end{figure}

In Fig.~\ref{fig:edit}, we further showcase examples of object movement and style transfer to demonstrate the potential of our method in more downstream editing tasks. For object relocation, we specify a target 2D coordinate under an up–down projection view and move the corresponding object point cloud to this destination in 3D space. Our rendering accurately reflects the relocation, producing images that maintain a high texture consistency with the original scene. For style transfer, we apply a style transformation to the input images, yielding rendered results that conform to the applied style modification of the inputs.

\section{Datasets}
We evaluate our reconstruction and rendering pipeline on testing scenes of the ScanNet dataset which are ``scene0758\_00'', ``scene0738\_\\00'', ``scene0708\_00'' and ScanNet++ which are ``3e6ceea56c'', ``4610b2\\104c'', ``6126572846'', ``a8f7f66985''.
We evaluate the instance segmentation results on annotated scenes from the ScanNet dataset that provide ground truth instance labels, which are: ``scene0107\_00'', ``scene0362\_00'', ``scene0409\_00'', ``scene0469\_00'', ``scene0616\_00''.
For all scenes in the ScanNet dataset, we resize each input image to a resolution of 624×468. For each scene, we selected a small number (10 to 20) of frames to cover the whole original scene content based on their degree of inter-frame overlap to establish the ``sparse-views'' setup for all experiments.
Similarly, for the scenes in the ScanNet++ dataset, all images are resized to 640×480, and 10 to 20 sparse views are sampled from each scene.
For novel view synthesis testing, we select 6-8 testing images for each scene.

\section{Limitations and Future Work}

Our method adopts a diffusion-based rendering pipeline for novel view synthesis, and while the current implementation has not yet achieved real-time rendering, it exhibits substantial optimization headroom. By incorporating acceleration techniques for diffusion inference, our approach holds strong potential to realize rapid and efficient rendering.
Our method enables both reconstruction and rendering without requiring any input camera parameters. To render from a given viewpoint, we introduce an additional camera pose alignment step to align the test view with our predicted coordinate system. It is important to note that this alignment process is sensitive to the pose estimation error of the feed-forward network. Therefore, improving the accuracy of camera pose prediction remains a direction for future work. 
Another direction for future work is to extend our framework to support more downstream tasks, such as complex scene-level manipulations.